\documentclass{article}

\usepackage[nonatbib,preprint]{neurips_2019}
\usepackage[utf8]{inputenc} % allow utf-8 input
\usepackage[T1]{fontenc}    % use 8-bit T1 fonts
\usepackage{hyperref}       % hyperlinks
\usepackage{url}            % simple URL typesetting
\usepackage{booktabs}       % professional-quality tables
\usepackage{amsfonts}       % blackboard math symbols
\usepackage{nicefrac}       % compact symbols for 1/2, etc.
\usepackage{microtype}      % microtypography
\usepackage{graphicx}
\usepackage{amsmath}

\title{Membership Model Inversion Attacks for Deep Networks}

\author{%
  Samyadeep Basu
    \\
  Department of Computer Science\\
  University of Maryland\\
  College Park, MD 20740 \\
  \texttt{sbasu12@cs.umd.edu} \\
   \And
   Rauf Izmailov \\
  Perspecta Labs \\
  Basking Ridge, NJ \\
  rizmailov@perspectalabs.com \\
   \AND
   Chris Mesterharm \\
  Perspecta Labs \\
  Basking Ridge, NJ \\
  jmesterharm@perspectalabs.com\\
}

\begin{document}

\maketitle

\begin{abstract}
  With the increasing adoption of AI, inherent
  security and privacy vulnerabilities for machine learning
  systems are being discovered. One
  such vulnerability makes it possible for an adversary to
  obtain private information about the types of instances used to
  train the targeted machine learning model. This so-called
  model inversion attack is based
  on sequential leveraging of classification scores towards
  obtaining
  high confidence representations for various classes. However, for
  deep networks, such procedures usually lead to unrecognizable
  representations that are useless for the adversary. In this paper, we
  introduce a more realistic definition of model inversion, where the adversary
  is aware of the general purpose of the attacked model
  (for instance,
  whether it is an OCR system or a facial recognition system),
  and the goal
  is to find realistic class representations within the
  corresponding lower-dimensional manifold
  (of, respectively, general symbols or general faces).
  To that end, we leverage properties of generative
  adversarial networks for constructing a connected lower-dimensional
  manifold, and demonstrate the efficiency of
  our model inversion attack that is carried out within that manifold.
\end{abstract}

\section{Introduction}
The last decade witnessed a rapid and significant progress in
developing and applying deep learning techniques. At the same time,
various concerns about security of deployed machine learning models
have increased as well. It has been already shown~\cite{42503} that
a small visually imperceptible perturbation of an image can cause a
deep neural network classify it incorrectly and with high
confidence. Besides these {\em adversarial attacks}, another type of
security threats in the form of {\em membership attacks} was
discovered recently: it was shown~\cite{Shokri} that an adversary
can identify if a given sample was used during the training phase of
the targeted machine learning model, thus endangering the privacy of
training data. Yet another type of privacy threat is {\em model
inversion} attack: it has been shown~\cite{Fredrikson2} that an
adversary can recover typical representations of specific target
classes by leveraging confidence scores of machine learning model.
So far, model inversion attacks have been mostly successful against
shallow machine linear models such as SVM and logistic regression;
however, for deep neural networks, model inversion attacks usually
return but unrecognizable solutions~\cite{Shokri} that are useless
for the adversary.

In this paper, we focus on the white-box model inversion attack
where the adversary has access to the model and attempts to generate
representative data similar to training instances (we call them
representative samples) of individual classes. In the most general
form of model inversion attack, there is no additional information
about the type of the problem that the targeted model is trained to
classify. This, however, appears to be an excessively strict
assumption since the adversary would have no way to interpret the
multi-dimensional solution vector that can be obtained as a result
of such attack. Instead, we assume the adversary has some general
knowledge of the problem, and we exploit that general information in
order to guide the search for representative samples. For instance,
the attacker might know that the input is an image with specific
dimensions that is used by the targeted machine learning application
such as optical character recognition (OCR) or facial recognition.
This is a rather realistic assumption since the attacker have to
know how to interpret the model inputs. The model inversion goal is
thus to learn specific details of the system such as what individual
symbols are used for the targeted OCR application or what faces can
be correctly identified by the targeted facial recognition security
application.

The direct search in the (very) high-dimensional input space (i.e.,
without {\em any} additional knowledge about the problem) is usually
an ill-posed problem~\cite{Mahendran_2015_CVPR}. However, with
additional information about the problem, we can constrain the
search to a smaller-dimensional manifold that likely contains the
training data. According to the manifold assumption
theory~\cite{Zhu}, many data sets belong to a group of disconnected
low-dimensional manifolds. However, if these manifolds can be linked
with each other, we can search in the resulting connected manifold.
Given that the adversary knows the general type of problem, an
appropriate generative adversarial network (GAN) can also be
generated, which would model a connected manifold structure
\cite{Khayatkhoei:2018:DML:3327757.3327836}. For example, the
manifold can be generated to attack an OCR system by creating a GAN
using characters from various languages and sets of symbols; for
attacking a facial recognition system, a diverse set of faces can be
downloaded from Internet. By connecting the output of the GAN to the
input of the model, various optimization techniques can be used to
search for manifold instances that maximize label confidence values.

\section{Method}
A Generative Adversarial Network, introduced by
\cite{NIPS2014_5423}, is a min-max game between two neural networks:
generator ($G_{\theta}$) and discriminator ($D_{\phi}$). The
generator $G_{\theta}$ takes random noise $z$ as input and generates
$G_{\theta}(z)$. The discriminator $D_{\phi}$ distinguishes between
real samples ($x$) and fake samples coming from $G_{\theta}$. The
objective function for the min-max game between $G_{\theta}$ and
$D_{\phi}$ is
\begin{equation}\label{gan}
  \min_{\theta} \max_{\phi} \mathbb{E}_{x
    \sim P(x)}[\log(D_{\phi}(x)] + \mathbb{E}_{z \sim P(z)}[1 -
  \log(D_{\phi}(G_{\theta}(z))].
\end{equation}
In \eqref{gan}, $P_{x}$ is the real data distribution, and $P_{z}$
is a noise distribution which is typically a uniform distribution or
a normal distribution.

Previous research has shown that real images have probability
distributions ($P_{x}$) on low-dimensional manifolds~\cite{Zhu}
embedded in a high-dimensional space. Intuitively, sufficiently
different images should belong to their own {\em disconnected}
manifolds without any paths of ``blended'' images between them.
However, in case of a GAN, the generator function maps from a
connected distribution space, like the uniform distribution, to all
possible outputs, which results in a {\em connected} output set of
instances. This is a typical drawback of GANs and various techniques
to partition the input into disjoint support sets have been used
\cite{Khayatkhoei:2018:DML:3327757.3327836} to address this issue.
However, our approach actually leverages this drawback in order to
search in the low-dimensional but connected latent space of the GAN
set instead of the high-dimensional space $P_{x}$ of all possible
images. Details of our approach are presented in Appendix.

A direct solution of model inversion problem without the use of the
GAN can be formulated as follows. Let $f_{\delta}$ be the target
neural network which is being attacked and $y$ be the one-hot
encoding vector representing the class, whose representative sample
needs to be recovered.  Let
\begin{equation}\label{eq: first}
  \hat{x} = \arg \min_{x} \ell (f_{\delta}(x), y) + \lambda R(x),
\end{equation}
where $\lambda$ is a regularization hyperparameter and $R(x)$ is a
regularization term which can be the $\ell_{p}$ norm of the image.
We modify this standard formulation in the following way to directly
search in the latent GAN space:
\begin{equation}\label{eq:test}
  \hat{z} = \arg \min_{z} \ell (f_{\delta}(G_{\theta}(z), y) + \lambda
  R(z).
\end{equation}
The final solution for the representative sample is given by
\begin{equation}
    \hat{x} = G_{\theta}(\hat{z}).
\end{equation}
Equation~\eqref{eq:test} can be solved using any quasi-Newton method
like gradient descent or adaptive learning rate methods like Adam.

\section{Experiments}
In our experiments, we assume general knowledge of the underlying
machine learning system. For example, if an attacker targets an OCR
system for determining what specific symbols the system was trained
on, then a dataset comprising various characters can be constructed
first and then used to train a GAN for creating a connected manifold
structure (the constrained search space) from which representative
samples of the target model will be recovered. We performed our
preliminary experiments on two datasets: (1) Numeric MNIST with
Arabic MNIST, and (2) Fashion MNIST Dataset
\cite{fashion}. In our experiments, we used a 2-layer feed-forward
neural network with ReLU activation for the target model; for
constructing the connected manifold, we used both standard GAN with
feed-forward networks and DCGAN \cite{dcgan}.

\subsection{Dataset 1: Numeric MNIST and Arabic MNIST}
In this case, the targeted deep neural network $f_{\delta}$ has been
trained with a subset of MNIST (namely 6 classes out of 10 classes).
We curated a dataset comprising of numeric MNIST (10 classes) and
Arabic MNIST (10 classes), which we used to train a GAN in order to
create the connected manifold for the search procedure according to
\eqref{eq:test}. The task is to identify representative samples from
the 6 classes with which $f_{\delta}$ was trained. Figure~1
%Figure~\ref{no-knowledge},
shows the results obtained using an optimization search in the full
image space. As expected, in this case, no representative samples
were found.
%Figure~\ref{knowledge1}
% and Figure~\ref{knowledge2}
In contrast to Figure~1, Figure~2 shows some of our results obtained
using our GAN-based technique: they can be clearly viewed by the
adversary as reasonably representative samples of the attacked
classes.
\begin{figure}
  \centering
  \fbox{ \includegraphics[width=0.3\textwidth]{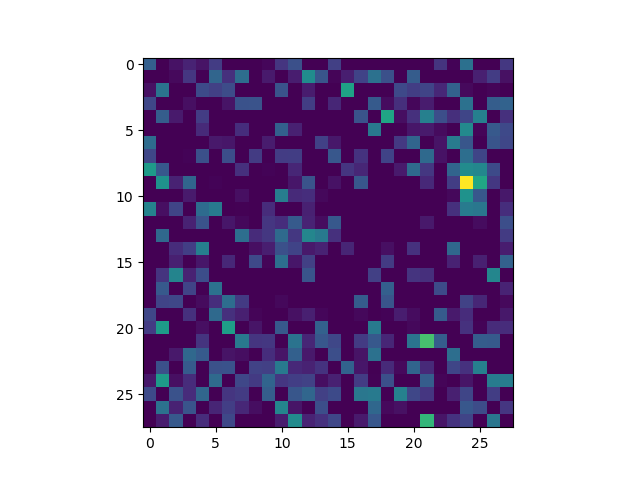}
    \includegraphics[width=0.3\textwidth]{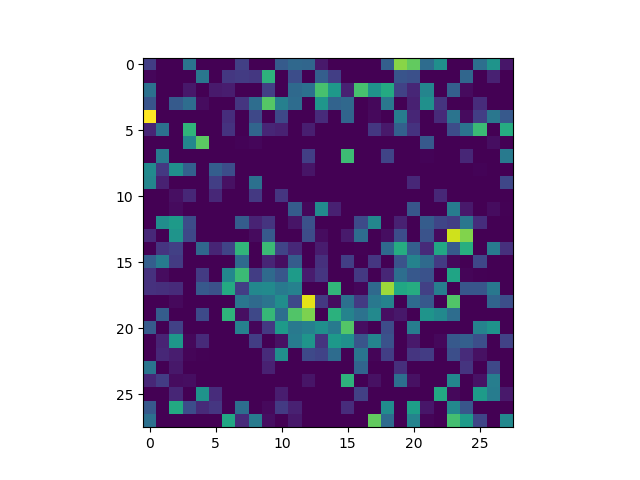}
    \includegraphics[width=0.3\textwidth]{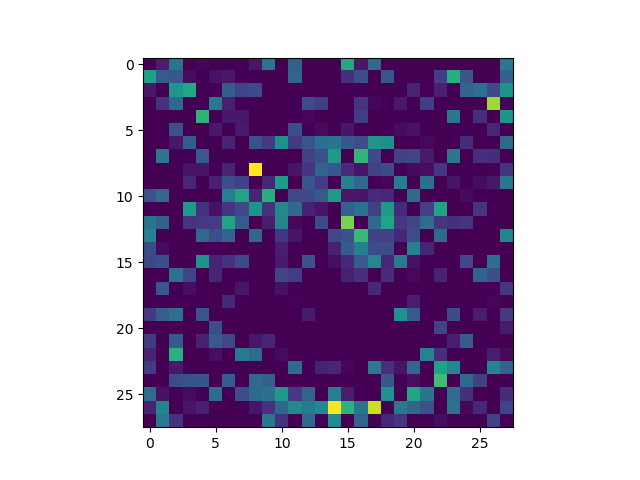} }
  \caption{Attack on MNIST classifier without background knowledge:
  (Left) Retrieval of class ``5'', (Middle) Retrieval of class ``6'',
  (Right) Retrieval of class ``9''.}
  \label{no-knowledge}
\end{figure}

\subsection{Dataset 2: Fashion MNIST}
In this set of experiments, we train a deep neural network with a
subset of Fashion MNIST \cite{fashion}, namely 5 classes out of 10;
more details on this dataset are presented in Appendix. We assume we
have knowledge about all different types of clothes and footwear
and, using the complete Fashion MNIST dataset, we construct the
connected manifold using a GAN. The attack objective is to identify
representative samples from the 5 classes on which the model was
trained. Figure~3 illustrates the successful recovery of its classes
using our approach.

\subsection{Effect of Regularization and High Order Terms}
We further used the $\ell_{p}$ norm regularization to improve the
quality of the images. With $\ell_{p}$ regularization, we solved
\eqref{eq:test} and evaluated our results for $p$ ranging from 1 to
6. However, in our experiments we observed that regularization did
not seem to affect the quality of retrieved samples in comparison to
no regularization. We also report the results pertaining to the
impact of high-order loss approximations in the Appendix.
\begin{figure}
  \centering
  \fbox{ \includegraphics[width=0.3\textwidth]{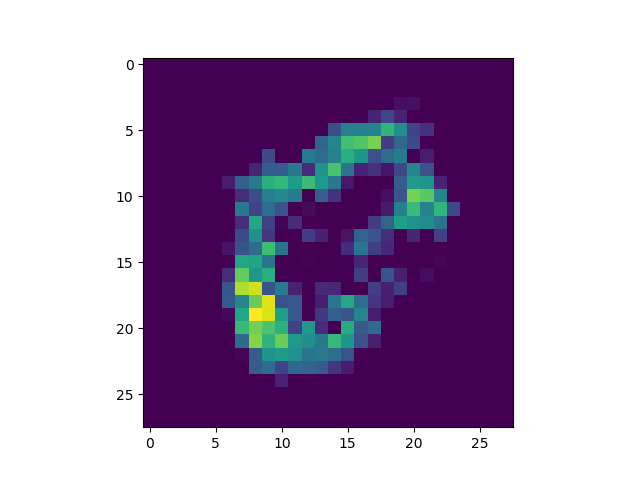}
    \includegraphics[width=0.3\textwidth]{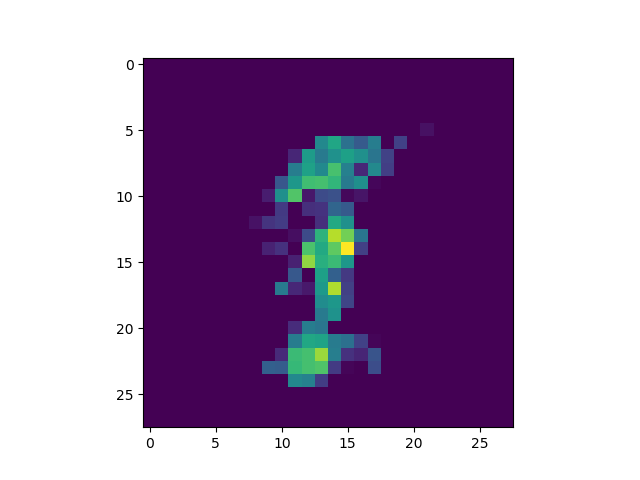}
    \includegraphics[width=0.3\textwidth]{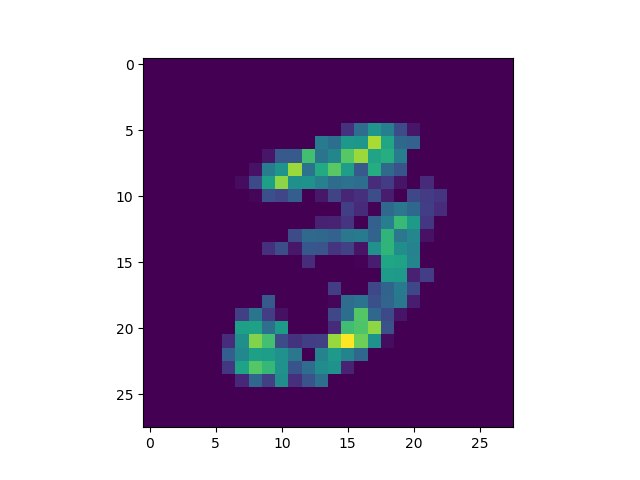} }
  \caption{Attack on MNIST classifier with background
knowledge:
  (Left) Retrieval of class ``0'', (Middle) Retrieval of class ``1'',
  (Right) Retrieval of class ``3''.
} \label{knowledge1}
\end{figure}

%\begin{figure}
%  \centering
%  \fbox{ \includegraphics[width=0.3\textwidth]{5_mnist.png}
%    \includegraphics[width=0.3\textwidth]{6_mnist.png}
%    \includegraphics[width=0.3\textwidth]{8_mnist.png} }
%  \caption{
%Attack on MNIST classifier with background knowledge:
%  (Left) Reconstruction of class ``5'', (Middle) reconstruction of class ``6'',
%  (Right) Reconstruction of class ``8''.
%    }
%    \label{knowledge2}
%\end{figure}

\begin{figure}
  \centering
  \fbox{ \includegraphics[width=0.3\textwidth]{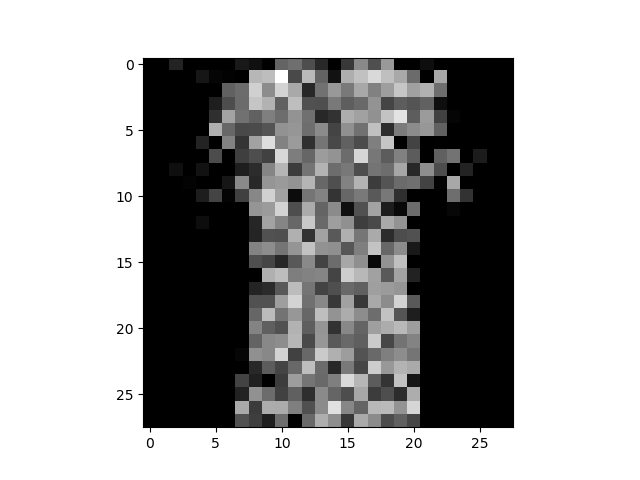}
  \includegraphics[width=0.3\textwidth]{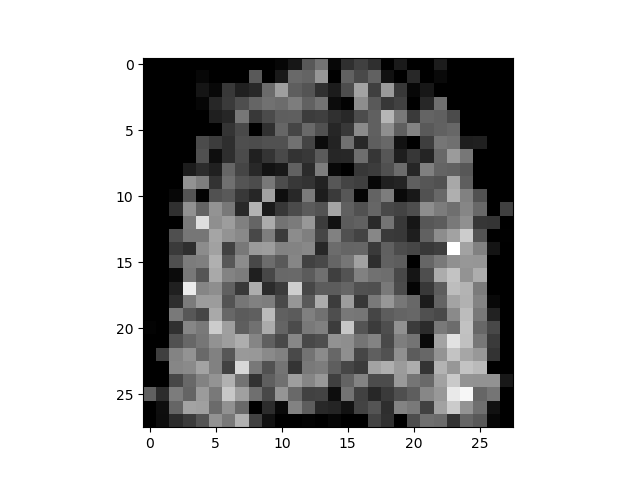} \includegraphics[width=0.3\textwidth]{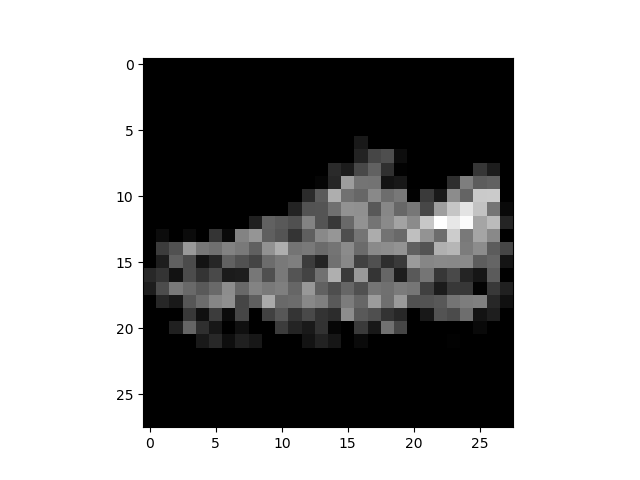}}
  \caption{Attack on Fashion MNIST classifier with background knowledge. (Left): Retrieval of Class ``T-shirts'';
  (Middle) Retrieval of class ``Coats''; (Right) Retrieval of class ``Sneakers''.}
\end{figure}

%\begin{figure}
 % \centering
 % \fbox{ \includegraphics[width=0.3\textwidth]{inital_data.PNG}  }
 % \caption{Samples from the curated dataset consisting of normal MNIST %digits and Arabic Digits}
 % \end{figure}

\section{Conclusion}

Without knowledge of the underlying machine learning system,
especially for deep neural networks, the model inversion problem is
ill-posed and the attack yields unrecognizable and unusable images
with high confidence values. However, with some natural knowledge
about the underlying target system, an attacker can use our
GAN-based approach for retrieving representative and recognizable
samples of individual classes.

We have also identified several promising research directions based
on our current work. One direction is to consider ways to exploit
the connected space of GANs.  Current research on GANs shows that it
is possible to generate interpretable images that are not clearly
related to the training data.  For example, GANs can be used to
generate realistic faces~\cite{stylegan} that were not part of the
training data. For applications such as facial recognition, this can
be useful to create a constrained GAN space to search. Even if a
specific face from the training data is not present in the data the
adversary uses to create the GAN, the resulting connected GAN space,
with its rich set of faces, might contain a face sufficiently close
to the face that was used to train the original machine learning
model.

Another research direction is to consider ways to develop a robust
defense against model inversion attacks {\em without} affecting the
model accuracy. This could be challenging since model inversion does
not deal with protecting any particular instance, so the defense
must protect all the representative images that are part of the
manifold used for training. Standard techniques such as differential
privacy \cite{Dwork:2006:DP:2097282.2097284} will be difficult to apply since they
focus on protecting specific finite sets of data.  Instead, it might
be useful to train a more complex classifier that has a larger set
of classes that can obscure the original ones. For example, a
security-based facial recognition system could classify a much
larger set of faces so that the faces that are actually relevant to
security verification are effectively hidden like a needle in a
haystack. The key problem here is to maintain adequate classifier
accuracy as the number of classes increases.

\section{Acknowledgments}

The authors are grateful for the support of NGA via contract number
NM0476-19-C-0007. The views, opinions and/or findings expressed are
those of the authors and should not be interpreted as representing
the official views or policies of the Department of Defense or the
U.S. Government. This paper has been approved for public release,
NGA \# 19-976.

\newpage
\bibliography{inversion}
\bibliographystyle{abbrv}
\section{Appendix} \label{eq: appendix}
\subsection{Impact of Second Order Terms on Model Inversion}
\subsubsection{Without general knowledge}
The higher order terms in the loss function \eqref{eq: first}, have
more information that can be used to improve the retrieved
representations. In case of ReLU networks, the obtained
representations do not look like original classes and are inherently
noisy in nature. The higher order terms, specifically the second
order term, has information about the geometry or curvature of the
loss surface, which might be informative to improve the
representations:
\begin{equation} \label{high_order}
    \hat{x} = \arg \min_{x} \ell (f_{\delta}(x), y) + \lambda R(x).
\end{equation}
We formulate a second order approximation of \eqref{high_order}
using Taylor's approximation:
\begin{equation}
    \min_{\delta x} \ell(x + \delta x) = \ell (x) +
    (\delta x)^{T}\nabla_{x} \ell (x) + \frac{1}{2} (\delta x)^{T} H_{x} (\delta
    x),
\end{equation}
where $x + \delta x$ is a stationary point for $\ell (x + \delta
x)$:
\begin{equation}
    \nabla_{\delta x} (\ell (x + \delta x)) = 0
\end{equation}
\begin{equation} \label{inverse_hvp}
    \delta x = - H_{x}^{-1} \nabla_{x} \ell(x)
\end{equation}
Although the computation of the Hessian will be expensive for
\eqref{inverse_hvp}, it can be solved efficiently using a conjugate
gradients which uses a Hessian-Vector formulation. HVP can be
efficiently solved using the Pearlmutter's formula
\cite{hvp_pearlmutter}. However, based on our experiments, the
second order term did not help in improving the representations, as
shown in Figure~4.
\begin{figure}
  \centering
  \fbox{ \includegraphics[width=0.3\textwidth]{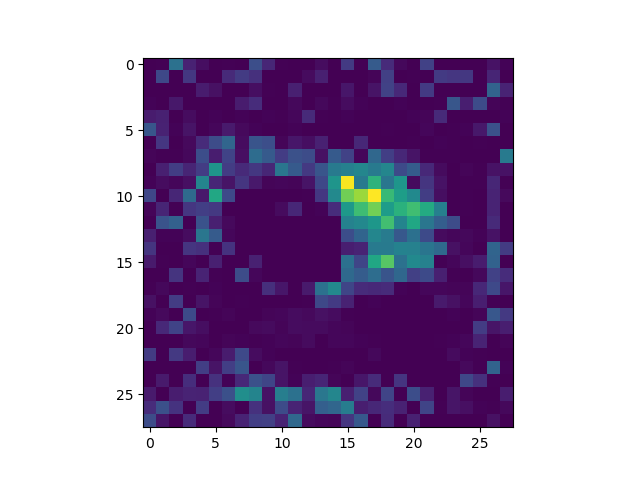} \includegraphics[width=0.3\textwidth]{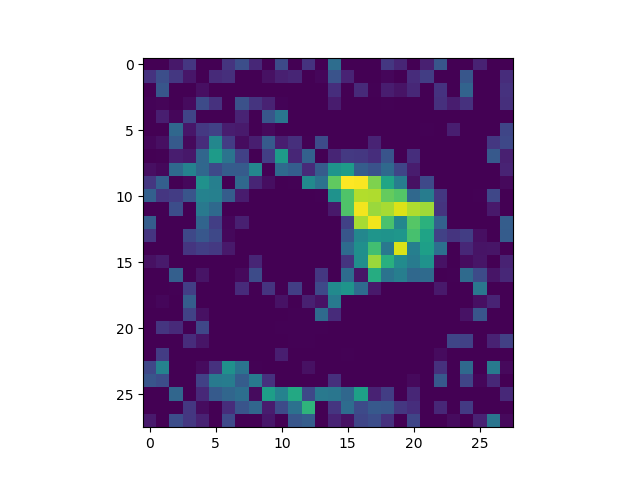}}
  \caption{High Order Loss approximations for retrieving class
  ``5'': (Left) high order loss approximation for Deep
  Sigmoid Networks; (Right) first order
  loss function.}
\end{figure}

\subsection{Lipschitzness of the Generator of the GAN}
In this section, we describe the connected manifold space of the GAN
and show how the connected structure helps in transition from one
image to another. Our analysis is similar to
\cite{agrawal2018towards}, where we assume that the the generator is
bounded. If the generator ($G$) is Lipschitz, we can show that there
is unused space in the latent space which corresponds to images
outside of the true manifold. If the generator is $\beta$-Lipschitz,
for any two points in the range of the generator and belonging to
two different and distinct manifolds, we have
\begin{equation} \label{lip1}
    \|G(z_{1}) - G(z_{2}) \| \leq \beta \| z_{1} - z_{2} \|.
\end{equation}
Essentially, \eqref{lip1} describes how fast the transition can take
place from one manifold to another. We denote the two manifolds as
$P$ and $Q$ and the true probability distribution as $P_{x}$. For
any two points $p \in P$ and $q \in Q$, The minimum distance between
two points in manifolds P and Q will be
\begin{equation} \label{lip2}
    \gamma = \min_{p \in P,q \in Q} \| p - q \|.
\end{equation}
Replacing \eqref{lip2} in \eqref{lip1}, we obtain the following
inequality:
\begin{equation} \label{lip3}
    \| z_{1} - z_{2}\| \geq \frac{\gamma}{\beta}
\end{equation}
This inequality shows that there will be a certain amount of unused
space when we are transitioning from one manifold to another. This
space could be thought of as a tunnel between different manifolds.
These transitioning tunnels help to recover images in the training
set, by searching in the latent space. Otherwise, if the latent
space is disconnected, then optimizing \eqref{eq:test} would not
result in the representative images, as there would be no possible
transition from one manifold to another in the latent space.

\subsection{Fashion MNIST}
Fashion MNIST dataset \cite{fashion} comprises of 10 classes of
different type of clothing, footwear, etc. We trained our target
deep network with a subset of the classes(5 out of 10) and then
performed a search in the latent space of the GAN which has been
trained on all the classes. Figure~5 shows the results for model
inversion without any assumption of background knowledge. Without
prior knowledge, the retrieved samples are extremely noisy and are
unrecognizable to the human eye. However, these noisy images produce
a high-confidence score for the target class.
\begin{figure}
  \centering
  \fbox{ \includegraphics[width=0.3\textwidth]{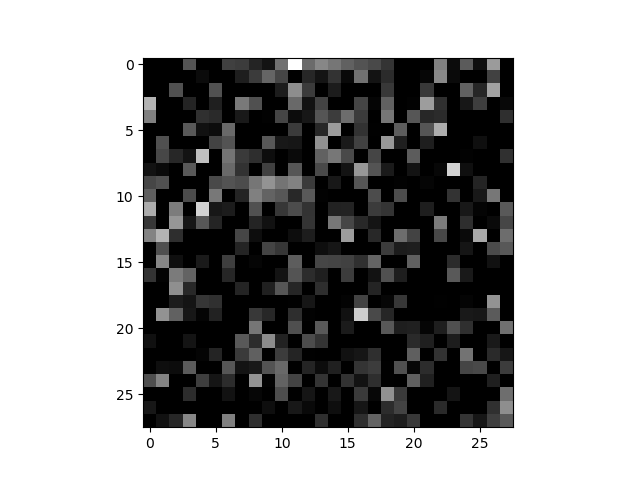} \includegraphics[width=0.3\textwidth]{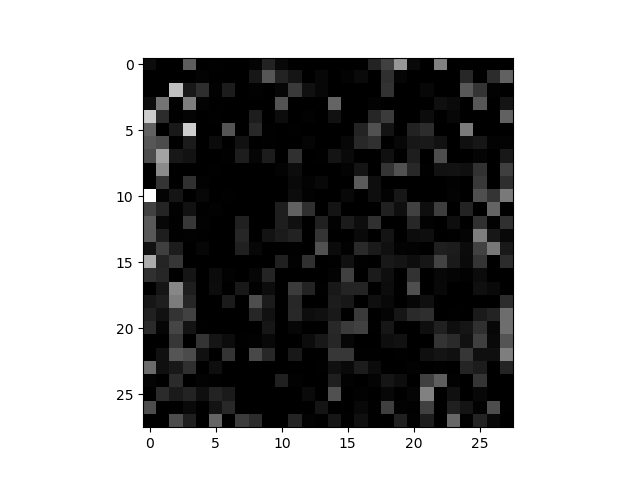} \includegraphics[width=0.3\textwidth]{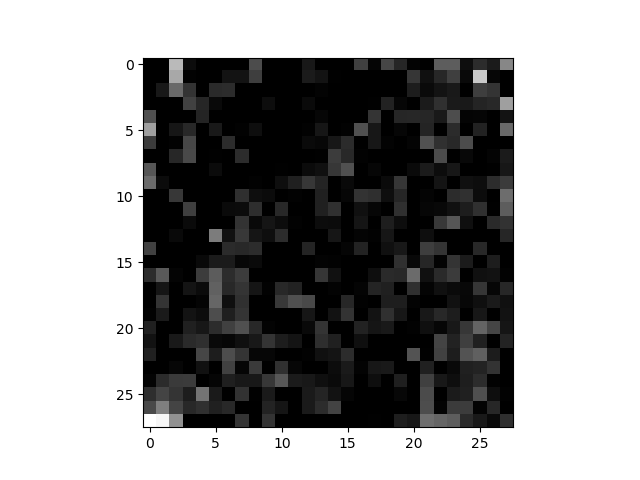}}
  \caption{Fashion MNIST with Deep ReLU networks: (Left): Retrieval of Class ``Trousers'';
  (Middle) Retrieval of class ``Sneakers''; (Right) Retrieval of class ``Bags''}
\end{figure}
Figure~3 and Figure~6 show the retrieval of representative training
samples with background knowledge about the system.

\begin{figure}
  \centering
  \fbox{ \includegraphics[width=0.3\textwidth]{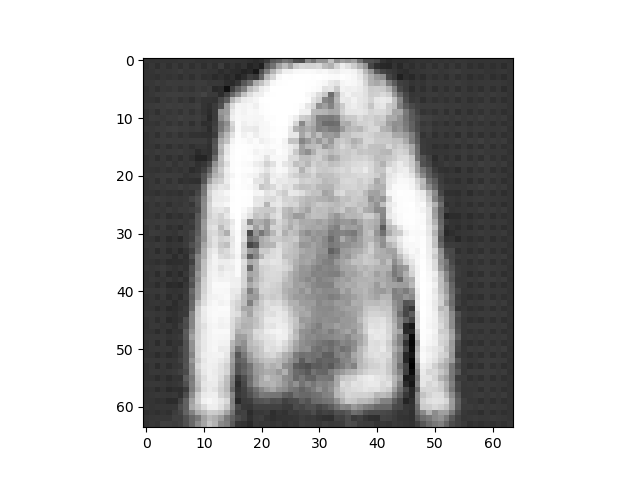} \includegraphics[width=0.3\textwidth]{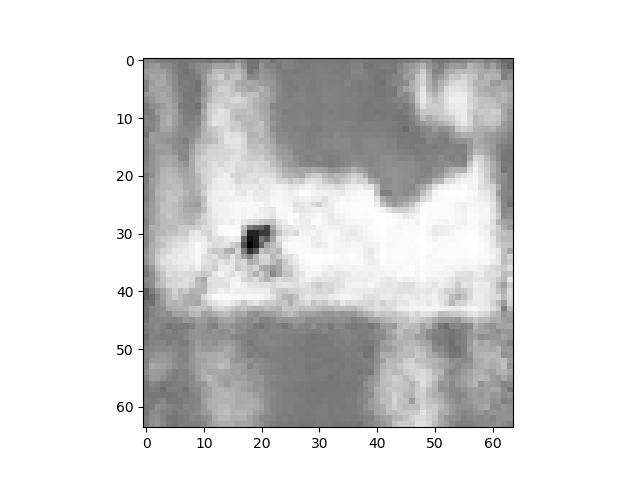} \includegraphics[width=0.3\textwidth]{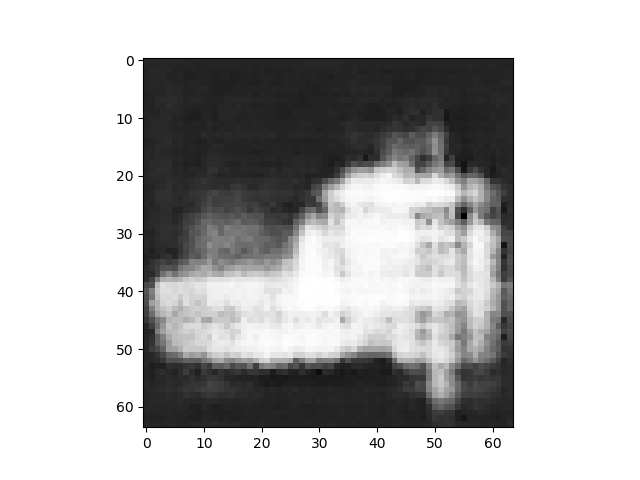}}
  \caption{Fashion MNIST with Deep ReLU networks with background knowledge [DCGAN](Left):
  Retrieval of Class ``Pullovers''; (Middle) Retrieval of class ``Bags''; (Right) Retrieval of class ``Sneakers''}
\end{figure}
\end{document}